\documentclass{article}

     \usepackage[nonatbib,preprint]{neurips_2019}

\usepackage[utf8]{inputenc} 
\usepackage[T1]{fontenc}    
\usepackage{hyperref}       
\usepackage{url}            
\usepackage{booktabs}       
\usepackage{amsfonts}       
\usepackage{nicefrac}       
\usepackage{microtype}      
\usepackage{amsmath}
\usepackage{subfig}
\usepackage{graphicx}
\usepackage{xspace}
\usepackage{booktabs}
\usepackage{tabularx}
\usepackage{multirow}
\usepackage{soul}
\usepackage{color}
\usepackage{threeparttable}
\usepackage{array}
\usepackage{multirow}
\usepackage{comment}

\newcolumntype{M}[1]{>{\centering\arraybackslash}m{#1}}
\newcommand{\addpic}[1]{\includegraphics[width=6em]{#1}}

\title{Do Explanations Reflect Decisions?  A Machine-centric Strategy to Quantify the Performance of Explainability Algorithms}

\author{Zhong Qiu Lin $^{1,2,3,*}$, Mohammad Javad Shafiee$^{1,2,3,*}$,  Stanislav Bochkarev $^{3}$\\ \textbf{Michael St. Jules$^{3}$, Xiao Yu Wang$^{3}$, Alexander Wong$^{1,2,3}$}\\
			$^{1}$ Vision \& Image Processing Group, Systems Design Engineering, University of Waterloo\\
			$^{2}$ Waterloo Artificial Intelligence Institute, Waterloo, ON\\
			$^{3}$ DarwinAI Corp., Waterloo, ON\\ 		
			$^{*}$ Equal Contribution \\ 					
	}

\begin{document}
\maketitle
\vspace{-0.15in}
\begin{abstract}
\vspace{-0.15in}
There has been a significant surge of interest recently in the research community around the concept of explainable artificial intelligence (XAI), where the goal is to produce an interpretation for a decision made by a machine learning algorithm.  Of particular interest is the interpretation of how deep neural networks make decisions, given the complexity and `black box' nature of such networks.  Given the infancy of the field, there has been very limited exploration into the assessment of the performance of explainability methods, with most evaluations centered around subjective visual interpretation of the produced interpretations.  In this study, we explore a more machine-centric strategy for quantifying the performance of explainability methods on deep neural networks via the notion of decision-making impact analysis.  More specifically, we quantify the importance of identified critical factors for a given decision made by a network based on the impact over network decisions and confidences in the absence of these critical factors.  For scenarios where we wish to study impact in directed erroneous decisions (e.g., under adversarial distractions), we additionally quantify importance of identified critical factors based on coverage of the adversarially impacted factors. We introduce two quantitative performance metrics: i) Impact Score, which assesses the percentage of critical factors with either strong confidence reduction impact or decision changing impact, and ii) Impact Coverage, which assesses the percentage coverage of adversarially impacted factors in the input.  A comprehensive analysis using this approach was conducted on several state-of-the-art explainability methods (LIME, SHAP, Expected Gradients, GSInquire) on a ResNet-50 deep convolutional neural network using a subset of ImageNet for the task of image classification. Experimental results show that, for both general and adversarial distraction scenarios, the critical regions identified by LIME within the tested images had the lowest impact on the decision-making process of the network ($\sim$38\%), with progressive increase in decision-making impact for SHAP ($\sim$44\%), Expected Gradients ($\sim$51\%), and GSInquire ($\sim$76\%).  A similar trend is observed in terms of impact coverage under adversarial distractions, with impact coverage being lowest for LIME and highest for GSInquire.  While by no means perfect, the hope is that the proposed machine-centric strategy helps push the conversation forward towards better metrics for evaluating explainability methods and improve trust in deep neural networks.

\end{abstract}
\vspace{-0.1in}
\section{Introduction}
\vspace{-0.15in}
\label{sec:intro}
The significant advances in deep learning~\cite{lecun2015deep}, in particular deep neural networks, has led to the rise in adoption across industry.  This has also led to a tremendous rise in research in the area of deep learning and its application for a wide variety of tasks, leading to state-of-the-art performance across various tasks such as visual, perception~\cite{szegedy2016rethinking,huang2017densely,redmon2017yolo9000}, speech recognition~\cite{amodei2016deep}, and natural language processing~\cite{young2018recent,gardner2018allennlp}.  However, as the proliferation of deep learning continues, there is now a growing interest as well as concern over how deep neural networks are making decisions, particularly for life-critical applications such as autonomous driving and clinical decision support.  Given the sheer complexity of deep neural networks and how information propagates through such networks to form a decision, deep learning has been often viewed as a `black box' machine learning method and very difficult to interpret and understand the decision-making process or the key factors involved in the decision.  This makes deep learning challenging to leverage particularly in regulated spaces where interpretability and transparency is a necessity (e.g., finance and healthcare).  Furthermore, this challenge of interpretability also makes it very difficult for machine learning engineers and scientists to understand biases and error scenarios of the trained network to improve upon, as well as situations where the network is deciding based on unintended patterns in the dataset~\cite{niven2019}.  This is particular critical given the recent rise of adversarial examples~\cite{szegedy,akhtar}, which are designed specifically to cause deep neural networks to make erroneous decisions and the understanding of how networks behave is very important to better devise ways to defend against them.  As such, the ability to explain the decision-making process of deep neural networks can be critical for enabling the development of improved, more dependable deep learning as well as enable the use of deep learning in a more trust-worthy manner in mission-critical scenarios.

Due to this critical need for increased transparency and interpretability in deep learning, there has been a considerable increase in research interest on explainability methods for interpreting the decision-making process of a deep neural network.  In the field of computer vision, such explainability methods typically manifest their interpretations of the decision-making process of visual perception neural networks in the form of visual saliency maps that highlight critical regions deemed by the method as important in making the decision.  While such visual interpretations aim to give new insights into the way deep neural networks make decisions, much of the evaluation around the visual interpretations produced by explainability methods have been largely subjective as such, ironically, is up to the interpretation of the human observer and thus difficult to judge whether these identified critical regions are in fact reflective of what the deep neural network is leveraging to make decisions.  While this current gap in the exploration of quantitative performance assessment of explainability methods in terms of their impact on decisions made by deep neural networks is understandable given how new this area of research is, this gap  hinders the level of human trust in not just the deep neural networks but also in the explainability methods themselves.  In fact, quantitative methods to assess the performance of explainability methods is critical to not only trust in decisions made but also in the choice of method for deployment and research development, especially since different explainability methods can produce drastically different explanations given the same input data and same model and so it is difficult to know if algorithmic extensions on such explainability approaches actually improves interpretability.

In this study, we explore a more machine-centric strategy for quantifying the performance of explainability methods on deep neural networks via the notion of decision-making impact analysis. More specifically, we introduce a new performance metric (which we will refer to as the \textbf{Impact Score}) for quantifying how well the critical factors identified by an explainability method reflects a given decision made by a network based on the impact over network decisions and confidences in the absence of these critical factors.  For scenarios where we wish to assess impact in directed erroneous decisions (e.g., under adversarial distractions), we introduce an additional performance metric (which we will refer to as \textbf{Impact Coverage}) for quantifying the coverage of the identified critical factors on the adversarial impacted factors. Based on these metrics, we conduct a comprehensive analysis of the performance of four different state-of-the-art methods from recent research literature on the task of image classification to study how such methods compare against each other in terms of how much impact the critical regions identified in the explanations produced by each method actually has on the  decision-making process under both general and adversarial scenarios. To the best of the authors' knowledge, this is the first systematic study to quantitatively assess the performance of several state-of-the-art explainability methods based on how impactful their explanations are to decisions made by a network under both normal and adversarial scenarios.

\section{Related Work}
\vspace{-0.15in}
\label{sec:RW}
The explainability methods in current research literature can generally divided into two main categories~\cite{tjoa2019survey}.  In the first category of explainability methods, which we will refer to as \textbf{proxy} strategies~\cite{lime,SHAP}, a deep neural network is approximated by a proxy model and the decision-making of the deep neural network is interpreted by querying the proxy model. In the second category, which we will refer to as \textbf{direct} strategies~\cite{sundararajan2017axiomatic,springenberg2014striving,selvaraju2016grad,smilkov2017smoothgrad,Expected-gradient,wong2018} the decision-making process of a deep neural network is mainly interpreted by studying the internal behaviour within a deep neural network directly and then surfacing that information as an explanation for the decision-making process of the network.
The most well-known of proxy methods is LIME~\cite{lime}, which takes advantage of a linear proxy model to approximate the behavioural of the targeted machine learning model and then interprets the original model based on the learnt proxy. Proxy approaches are considered as 'black box' approaches where the explainability method does not have direct access to the inner workings of the network and the proxy model approximates it given the input and the output to the network.
On the other hand, direct explainability algorithms are usually considered as `white box' methods as they require access to the inner workings of a deep neural network such as gradients and activations at different layers for a given input to identify the key factors within the input that is critical to  the decision-making process. For example, by leveraging information about gradients, it is possible to  quantify how much change in the input data would turn the decision of the network to another output and as such measure the importance of each input in the decision-making process. Notable gradient-based direct explainability approaches include Integrated Gradient ~\cite{sundararajan2017axiomatic}, Guided Backpropagation~\cite{springenberg2014striving}, Guided GradCAM~\cite{selvaraju2016grad}, SmoothGrad~\cite{smilkov2017smoothgrad} and Expected Gradients~\cite{Expected-gradient}.
\vspace{-0.15in}
\section{Quantifying Explainability}
\label{sec:QE}
\vspace{-0.15in}
Much of research literature around explainability, particularly for visual perception tasks such as image classification, has revolved around subjective visual interpretation of the explanations produced by the explainability method.  This usually takes on the form of visual saliency maps, where salient regions in the map produced using the explainability method of choice are considered as critical regions influencing the decision made by a network.  However, due to the purely qualitative nature of such visual assessments, it is very challenging to get a good sense as to how well an explainability method is performing, how useful or meaningful the provide explanation is relative to its influence over the network's decision and its associated confidence, and more importantly how well it performs compared to other explainability methods.  As such, this can limit progress in the field of explainable artificial intelligence since there is no method of benchmarking based on subjective visual assessment.

More recently, there have been explorations into human-centric strategies for quantifying explainability performance in the case of visual perception, where the visual saliency map produced using a given explainability method for a given image is compared with a visual attention maps created based on gaze information collected from human subjects~\cite{lai2019human}.  While such an approach is a step towards quantification of explanations produced by explainability methods,  one of the biggest limitations of such an approach is the underlying assumption that a deep neural network makes decisions in a similar manner as human subjects, which is often not true.  As such, this human-centric approach to quantifying explainability performance provides very little insight on the actual driving factors of the decision-making process of deep neural networks.  Furthermore, this approach requires considerable human gaze information to be collected, which is simply impractical for most real-world scenarios.

To address the limitations of human-centric strategies for quantifying the performance of explainability methods, we take a drastically different direction by instead exploring a more machine-centric strategy where we quantify performance based on the decision-making behaviour of the network itself.  More specifically, we aim to quantify the performance of explainability methods on deep neural networks via the notion of decision-making impact analysis, where we instead study the quantitative impact of critical factors identified by an explainability method for a given decision made by a network based on the changes in decisions and associated confidences in the decisions of the network itself.

In the below sections, we will first define a performance metric for quantifying the impact of critical factors identified by an explainability method on decisions and the confidence in those decisions as made by a given deep neural network.  Next, we introduce an additional performance metric for directed erroneous decision scenarios based around the concept of impact coverage.

\subsection{Assessing Impact on Decisions}
\vspace{-0.15in}
In order to be facilitate for the quantitative assessment of the performance of a given explainability method, the first step is to first define and formulate a performance metric for performing such an assessment.  Motivated towards taking a machine-centric strategy to quantitative performance assessment of a given explainability method on a particular deep neural network, we aim to develop metrics that quantify the importance of critical factors identified by the explainability method for a given decision made by a network based on the impact these factors have over network decisions and the associated confidences.
We consider the critical factors $c$ identified by an explainability method $M$ to be important to a decision $y$ made by a deep neural network $N$ for a given input $x$ if either of the following conditions are met:
\begin{itemize}
    \item \textbf{Decision-level impact}:  The decision made by the deep neural network changes in the absence of the critical factors.
    \item \textbf{Confidence-level impact}:  The confidence of the deep neural network in its decision $z$ changes by $\tau$\% in the absence of the critical factors.
\end{itemize}

The motivation behind this definition of importance for critical factors as identified by a given explainability method is based on the idea that, if the critical factors are indeed crucial to the decision-making process of the deep neural network, then the absence of these critical factors in the given input will have such an impact that the network  behaves in a way that it would either be significantly less confident in its current decision, or so unconfident in its decision that its confidence in another decision is higher and thus leads the network to make a different decision all together.

In this study, we formulate the performance metric $I$, which we will refer to as the Impact Score, as follows.   Let the relationships between the critical factors $c$, explainability method $M$, the input  $x$, the decision $y$, confidence in the decision $z$, and the network $N$ be expressed by the following equations:
\begin{align}
\{y,z\}=N(x),
\end{align}
\begin{align}
c=M(x,N),
\end{align}
\noindent where $c \in x$.
Based on this, we can define the input in absence of $c$ as identified by $M$ as,
\begin{align}
x' = x - c,
\end{align}
and the decision given $x'$ as input into $N$ as,
\begin{align}
\{y',z'\}=N(x').
\end{align}

Therefore, in the general scenario, based on the conditions defined above that the critical factors $c$ for a given input $x$ as identified by $M$ must meet to be deemed as important, we can define the Impact Score $I$ across a set of $n$ inputs $X=\{x_1,x_2,\ldots,x_n\}$ as:
\begin{align}
I = \frac{1}{n}\sum_{i = 1}^{n}\left(\left(y'_i \neq y_i\right) \vee \left(z'_i \leq \tau z_i\right)\right).
\end{align}

\noindent where $i$ denotes the $i^{\rm{th}}$ input. In this study, we set $\tau=0.5$ to indicate that the network has lost half of the confidence it had on its original decision.  Finally, we also introduce a stricter variant of the above Impact Score, denoted by $I_{strict}$ where we only consider decision-level impact:
\begin{align}
I_{strict} = \frac{1}{n}\sum_{i = 1}^{n}\left(y'_i \neq y_i\right).
\end{align}

\textbf{Impact Coverage.}  In the scenario where we wish to study impact in directed erroneous decisions (e.g., decisions made under the influence of adversarial examples), we introduce an additional approach to quantitatively assessing performance of the different explainability methods since the critical factors that the network leverages to make a decision are largely known a priori to the evaluation (e.g., in the case of an adversarial patch, the critical region that is important to the decision-making process is the adversarial patch itself)  More specifically, we can further quantify importance of the identified critical factors $c$ based the amount of coverage of the adversarially impacted factors in $x$ by the critical factors $c$.

Let us define the Impact Coverage metric $I_{coverage}$ across a set of $n$ inputs $X=\{x_1,x_2,\ldots,x_n\}$ based on the intersection-over-union between the adversarially impacted factors and the critical factors across the given set of inputs:
\begin{align}
I_{coverage} = \frac{1}{n}\sum_{i = 1}^{n}\frac{\left(a_i  \vee c_i\right)}{\left(a_i  \cup c_i\right)}.
\end{align}
\noindent where $a_i$ is the adversarially impacted factors in input $x_i$. As such, the Impact Coverage metric is designed to be high when heavy overlapping between the identified critical factors and the adversarially impacted factors to reward strong alignment between the explanation produced by the explainability method and the actual factors impacting decision.

\section{Experimental Setup}
\label{sec:exp}
\vspace{-0.15in}
The conducted experiments and the explainability methods used in this study are described below.
\vspace{-0.1in}
\subsection{Experiment 1: General Scenario}
\begin{figure}[t]
    \centering
    \includegraphics[width=0.8\textwidth]{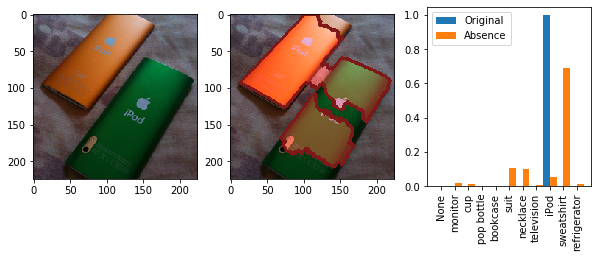}
    \caption{Example of a decision change due to absence of critical regions in the decision-making process. (left) original image; (center) identified critical region; (right) prediction confidences for decisions made with original image and with the absence of critical regions. The absence of critical regions led to a change in decision, which means the explanation reflects impact on the decision.}
    \label{fig:mask}
    \vspace{-0.15in}
\end{figure}

\vspace{-0.15in}
For the first experiment, we quantitatively evaluate the performance of several state-of-the-art explainability methods using the two variants of Impact Score (i.e., $I$ and $I_{strict}$) for each explainability method $M$ using a ResNet-50 deep convolutional neural network designed for the task of image classification as the reference network $N$.  A subset of the ImageNet ~\cite{russakovsky2015imagenet} dataset is leveraged as input $X$.  More specifically, we leveraged a subset of 410 different images from the ImageNet dataset, all of which had correct classifications for consistency purposes.  As such, this experiment tasks the different explainability methods to identify critical regions within a natural image that is important to the class prediction made by the network, such that in the absence of such critical regions the confidence by the network in the predicted class is either significant reduced or results in an altogether different class prediction.  An example of a decision change that resulted from the absence of critical regions identified by an explainability method during the decision-making process is known in Fig.~\ref{fig:mask}.  The purpose of this first experiment is the quantitatively evaluate explainability performance under a more general scenario where decisions are made on untampered data inputs and decisions are made by the network on such data inputs, and is representative of the general use case.

\vspace{-0.1in}
\subsection{Experiment 2: Adversarial Distraction}
\begin{figure}
    \centering
    \includegraphics[width=0.8\textwidth]{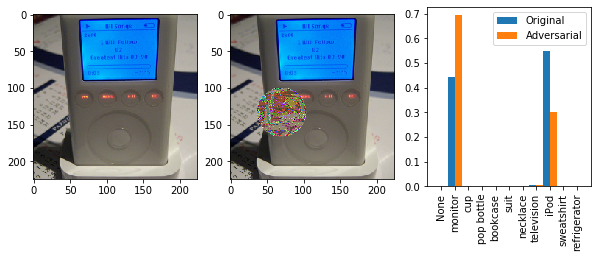}
    \caption{Example of a directed erroneous decision due to adversarially impacted area. (left) original untampered image, (center) tampered image with an adversarial patch, (right) prediction confidences of decisions made with untampered image and adversarially tampered image. The adversarial patch led to a change in decision.}
    \vspace{-0.15in}
    \label{fig:patch}
\end{figure}

\vspace{-0.15in}
For the second experiment, we quantitatively evaluate the performance of several state-of-the-art explainability methods using the two variants of Impact Score (i.e., $I$ and $I_{strict}$), as well as $I_{coverage}$ for each explainability method $M$ in the presence of visual 'distractions' in the form of adversarial patches to better study the impact in directed erroneous decisions.  More specifically, we leverage the adversarial patches from the work of Brown et al.~\cite{brown2017adversarial}. For generating the adversarial patch, we fix the reference network $N$ aforementioned in Experiment 1, and apply adversarial training for the same subset of the ImageNet ~\cite{russakovsky2015imagenet} dataset as Experiment 1. Later, we randomly (e.g. random translation and random rotation of the patch) overlay the resulting adversarial patches on the same subset of images with different patch scales ranging from $0.3$ to $0.7$.  An example of a directed erroneous decision due to adversarially impacted area is known in Fig.~\ref{fig:patch}.  We compute $I$, $I_{strict}$, and $I_{coverage}$ for each patch scale over the test images, of which the prediction classes change to the adversarially targeted classes. With the adversarial patch being the control variable, the critical region that is important to the decision-making process is largely known a prior to be the adversarial patch itself, and as such $I_{coverage}$ provides an additional quantitative indicator for the ability of the explainability method to identify such adversarially impacted areas within the images that has a direct impact in the decisions made by the deep neural network.

\vspace{-0.1in}
\subsection{Explainability Methods Under Study}
\label{sec:EM}
\vspace{-0.15in}
In this study, the proposed Impact Score and Impact Coverage is leveraged to perform a comprehensive analysis of on several state-of-the-art explainability methods in research literature.  More specifically, the methods under study are: i) LIME~\cite{lime}, ii) SHAP~\cite{SHAP}, iii) Expected Gradients~\cite{Expected-gradient}, and iv) GSInquire~\cite{wong2018}.  These methods were selected as they represent a good coverage of both popular and state-of-the-art methods from both the proxy and direct categories of explainability methods.

\vspace{-0.2in}
\section{Experimental Results}
\label{sec:ER}
\vspace{-0.15in}
The experimental results for the two experiments conducted in this study is presented below.

\begin{figure}[t]
    \centering
    \caption{Example images, the corresponding critical regions identified by tested explainability methods, and prediction confidences with and in absence of the identified critical regions.  }
    \begin{tabular}{cM{30mm}M{30mm}M{30mm}M{30mm}}
        \toprule

        & LIME~\cite{lime}
        & SHAP~\cite{SHAP}
        & Expected Gradients~\cite{Expected-gradient}
        & GSInquire~\cite{wong2018} \\

        \midrule
        & \addpic{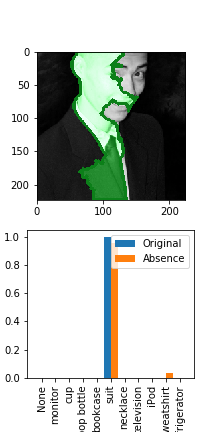}
        & \addpic{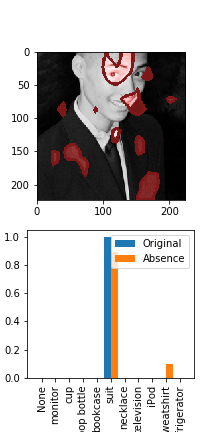}
        & \addpic{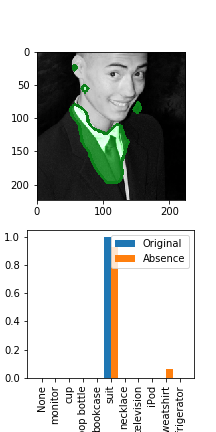}
        & \addpic{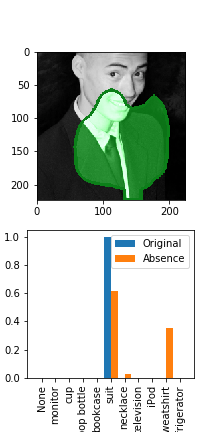} \\

        & \addpic{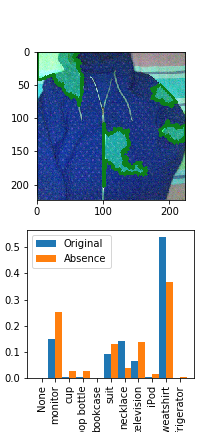}
        & \addpic{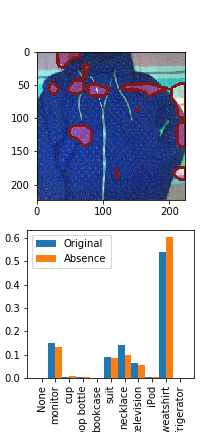}
        & \addpic{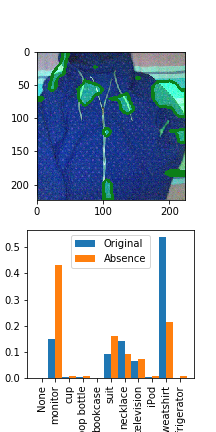}
        & \addpic{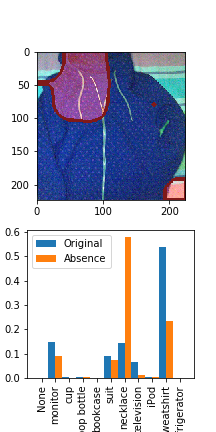} \\

        & \addpic{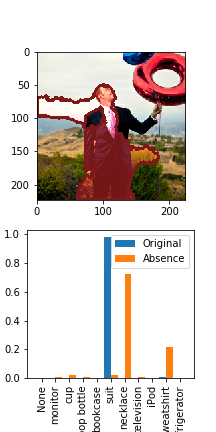}
        & \addpic{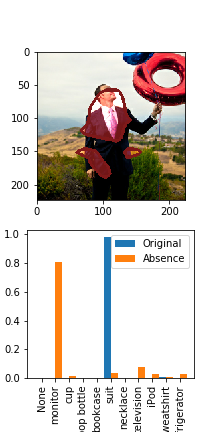}
        & \addpic{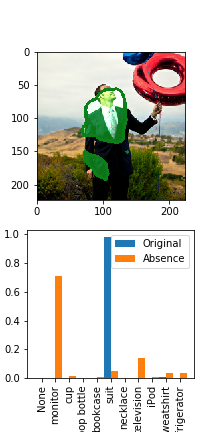}
        & \addpic{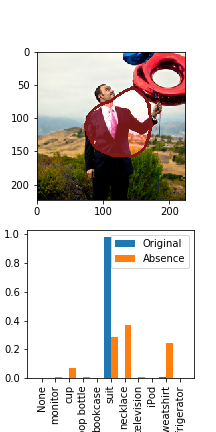} \\

    \end{tabular}
    \label{fig:experiment1a}
    \vspace{-0.2in}
\end{figure}

\textbf{Experiment 1:} The quantitative performance of the four tested explainability methods as determined by the proposed Impact Scores in the first experiment is shown in Table 1.  A number of interesting observations can be made.  First, it can be observed that LIME achieved the lowest $I$ and $I_{strict}$ scores, thus indicating that the critical regions identified by LIME had the lowest impact on the actual decision-making process of the network in identifying the class for a given image when compared to the other tested methods, with a difference in $I$ and $I_{strict}$ between  SHAP and LIME of over 6\% and over 5\%, respectively.  Second, it can be observed that there is a progressive increase in decision-making impact from SHAP to Expected Gradients, with a significant absolute increase in $I$ and $I_{Strict}$ by over 7\% and over 7.5\%, respectively.  While both SHAP and Expected Gradients approximate Shapley values, this significant improvement achieved by Expected Gradients over SHAP can be attributed to the incorporation of ideas behind three of most recent state-of-the-art concepts in explainability (SHAP, Integrated Gradients~\cite{sundararajan2017axiomatic}, and SmoothGrad~\cite{smilkov2017smoothgrad}) within a common expected value formulation, leading to the identification of more impactful critical regions.  Third, it can be observed that GSInquire achieved the highest $I$ and $I_{strict}$ scores amongst the tested methods, achieving a significant absolute increase of close to 25\% and close to 3\% in $I$ and $I_{strict}$, respectively, when compared to Expected Gradients.  What's interesting about this observation is the fact that the improvements of GSInquire for $I$ is significantly higher than the improvements for $I_{strict}$, which indicates that a much larger number of tested images experienced a significant confidence-level impact in the absence of the critical regions identified by GSInquire when compared to those identified by the other methods, while the improvements in decision-level impact is significant but less drastic.  An example image, the critical regions identified by tested explainability methods, and the prediction confidences with and in absence of the identified critical regions are shown in Fig.~\ref{fig:experiment1a}.  It can be observed that for one of the example images where both Expected Gradients and GSInquire identified decision-impacting critical regions while SHAP and LIME did not (middle row), the absence of the critical regions that SHAP identified not only did not lead to a decision change by the network, but instead led to an \textbf{increase} in prediction confidence for the original decision and as such illustrative of an explanation that does the reflect the decision-making process of the network.  Furthermore, as illustrated by the example image in the first row, no explainability method is perfect and the critical regions identified may not have decision-level impact (In this example, while decisions did not change in the absence of identified critical regions, the critical regions identified by GSInquire led to the highest prediction confidence change amongst the test methods).

\begin{table}[h]
    \centering
    \label{table:impact}
    \caption{Performance of tested explainability methods based on impact on network decisions.}
    \begin{tabular}{l|ccc}
        Method        & \multicolumn{1}{c}{$I$} & $I_{Strict}$  \\ \hline\hline
        LIME ~\cite{lime}         & 38.05\%                    & 35.12\%                              \\
        SHAP \cite{SHAP} & 44.15\%                    & 40.24\%                              \\
        Expected Gradients ~\cite{Expected-gradient}      & 51.22\%                           &      47.80\%                                     \\
        GSInquire~\cite{wong2018}     & 76.10\%                    & 50.73\%
    \end{tabular}
\end{table}

\begin{figure}[h]
    \centering
    \caption{Example adversarially modified erroneous images via adversarial patches at different scales, and the corresponding critical regions identified by tested explainability methods as being important to the decision made by the network. }
    \begin{tabular}{cM{20mm}M{20mm}M{20mm}M{20mm}M{20mm}}
        \toprule
        Patch Scale
        & Ground Truth / Adversarial Label
        & LIME~\cite{lime}
        & SHAP~\cite{SHAP}
        & Expected Gradients~\cite{Expected-gradient}
        & GSInquire~\cite{wong2018} \\

        \midrule
        0.30
        & Television / Monitor
        & \addpic{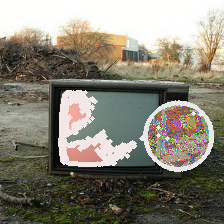}
        & \addpic{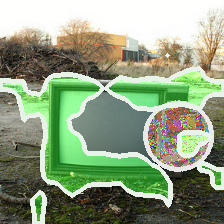}
        & \addpic{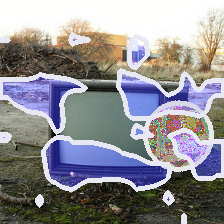}
        & \addpic{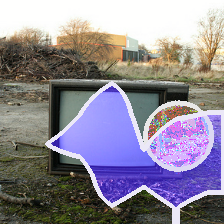} \\
        0.40
        & Suit / Cup
        & \addpic{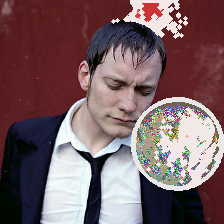}
        & \addpic{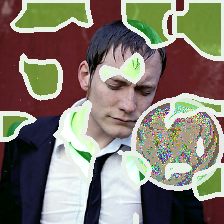}
        & \addpic{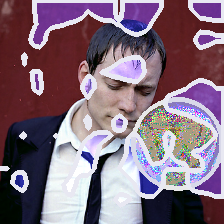}
        & \addpic{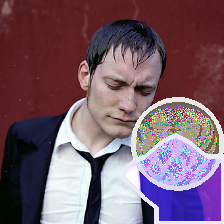} \\
        0.50
        & Necklace / Cup
        & \addpic{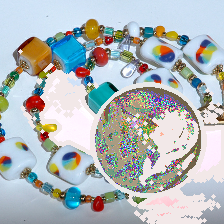}
        & \addpic{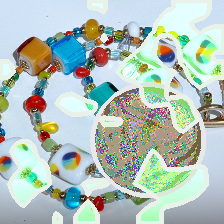}
        & \addpic{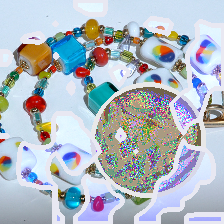}
        & \addpic{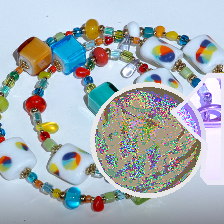} \\
        0.60
        & Sweatshirt / Monitor
        & \addpic{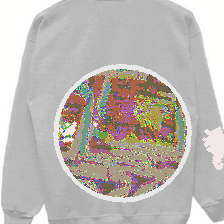}
        & \addpic{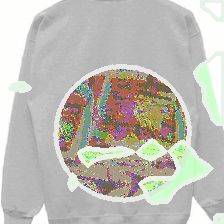}
        & \addpic{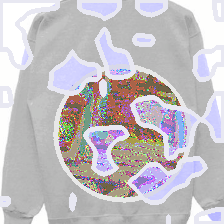}
        & \addpic{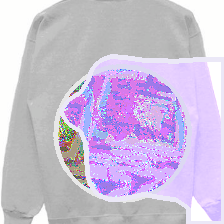} \\
        0.70
        & Cup / Necklace
        & \addpic{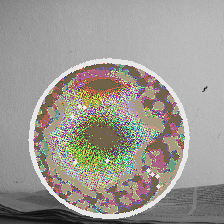}
        & \addpic{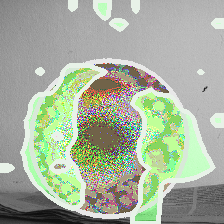}
        & \addpic{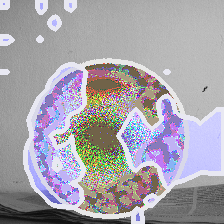}
        & \addpic{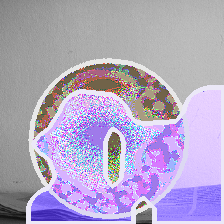} \\
    \end{tabular}
\vspace{-0.25in}
    \label{tab:experiment2}
\end{figure}

\textbf{Experiment 2:} The quantitative performance of the four tested explainability methods as determined by the proposed Impact Score and Impact Coverage in the second experiment is shown in Table 2.  A number of interesting observations can be made.  First, it can be observed that LIME achieved the lowest $I$, $I_{strict}$, and $I_{coverage}$ scores across all adversarial patch scales, thus indicating that the critical regions identified by LIME have the lowest impact as well as coverage of the adversarially impacted areas in the test images amongst the tested methods.  Second, it can be observed that both SHAP and Expected Gradients had similar $I$, $I_{strict}$, and $I_{coverage}$ scores, while GSInquire had significantly higher $I$, $I_{strict}$, and $I_{coverage}$ scores than both SHAP and Expected Gradients across all adversarial patch scales.  Example adversarially modified erroneous images via adversarial patches and the corresponding critical regions identified by tested explainability methods as being important to the decision made by the network are shown in Figure~\ref{tab:experiment2}.  It can be observed that both Expected Gradients and GSInquire were both able to better identify more adversarially impacted areas, with GSInquire achieving the best identification coverage for the adversarially impacted areas.

\begin{table}[]
\centering
\caption{Performance of tested explainability methods at different adversarial patch scales}
\resizebox{\textwidth}{!}{
\begin{tabular}{c|rrr|rrr|rrr|rrr}
\multicolumn{1}{l|}{\multirow{2}{*}{Scale}}
    & \multicolumn{3}{c|}{LIME~\cite{lime}}
    & \multicolumn{3}{c|}{SHAP~\cite{SHAP}}
    & \multicolumn{3}{c|}{Expected Gradient~\cite{Expected-gradient}}
    & \multicolumn{3}{c}{GSInquire~\cite{wong2018}}
    \\
\multicolumn{1}{l|}{}                       & \multicolumn{1}{c}{$I_{coverage}$} & \multicolumn{1}{c}{$I$} & \multicolumn{1}{c|}{$I_{strict}$} & \multicolumn{1}{c}{$I_{coverage}$} & \multicolumn{1}{c}{$I$} & \multicolumn{1}{c|}{$I_{strict}$} & \multicolumn{1}{c}{$I_{coverage}$} & \multicolumn{1}{c}{$I$} & \multicolumn{1}{c|}{$I_{strict}$} & \multicolumn{1}{c}{$I_{coverage}$} & \multicolumn{1}{c}{$I$} & \multicolumn{1}{c}{$I_{strict}$} \\ \hline
0.3                                         & 0.64\%                   & 9.70\%                  & 9.80\%                          & 3.53\%                   & 40.41\%                 & 41.32\%                         & 2.57\%                   & 36.00\%                 & 36.80\%                         & 13.90\%                  & 66.90\%                    & 68.00\%                        \\
0.4                                         & 1.53\%                   & 9.90\%                  & 10.00\%                         & 3.33\%                   & 36.73\%                 & 37.54\%                         & 2.31\%                   & 35.00\%                 & 35.40\%                         & 19.24\%                  & 64.50\%                    & 65.80\%                        \\
0.5                                         & 0.67\%                   & 8.70\%                  & 8.80\%                          & 3.08\%                   & 36.28\%                 & 36.62\%                         & 2.09\%                   & 39.20\%                 & 39.40\%                         & 20.02\%                  & 66.90\%                    & 67.80\%                        \\
0.6                                         & 0.37\%                   & 10.50\%                 & 10.60\%                         & 3.04\%                   & 38.20\%                 & 38.78\%                         & 1.88\%                   & 39.00\%                 & 39.40\%                         & 19.09\%                  & 67.20\%                    & 67.90\%                        \\
0.7                                         & 0.41\%                   & 10.80\%                 & 10.80\%                         & 2.87\%                    & 43.16\%                 & 43.61\%                         & 1.80\%                   & 42.80\%                 & 43.20\%                         & 17.29\%                  & 68.90\%                    & 69.70\%
\end{tabular}
}
\end{table}

\vspace{-0.15in}
\section{Conclusions}
\label{sec:conclusions}
\vspace{-0.15in}
In this study, we explored a more machine-centric strategy for quantifying the performance of explainability methods
on deep convolutional neural networks by quantifying the importance of critical factors identified by an explainability method for a given decision made by a network.  This is accomplished by studying the impact of identified factors on the decision and the confidence in the decision, and additionally the coverage of adversarially impacted factors in the directed erroneous decision scenario.  A comprehensive analysis using this approach showed that, in the case of visual perception tasks such as image classification, some of the most popular and widely-used methods such as LIME and SHAP may produce explanations that may not be as reflective as expected of what the deep neural network is leveraging to make decisions. Newer methods such as Expected Gradients and GSInquire performed significantly better in general scenarios, and GSInquire performing significantly better in adversarially distraction scenarios as well, though there is significant room for improvement and thus illustrating the importance of such quantitative metrics for benchmarking methods to better understand where our current approaches stand and where we can improve.   While by no means perfect, the hope is that the proposed machine-centric strategy helps push the conversation forward towards better metrics for evaluating explainability methods in a manner that gives insights to guide network error mitigation as well as improve trust in deep neural networks.  Future work involves studying the quantitative performance of explainability methods under different use cases such as speech recognition and natural language processing tasks, as well as the extension of the proposed Impact Score to incorporate a wider range of factors for more thorough quantitative assessment.

\medskip
\small

{\small
\bibliographystyle{plain}
\bibliography{refs}
}

\end{document}